\documentclass[conference]{IEEEtran}
\usepackage{times}

\usepackage[numbers]{natbib}
\usepackage{multicol}
\usepackage[bookmarks=true]{hyperref}
\usepackage{algorithm}
\usepackage{algpseudocode}
\usepackage{amsmath}
\usepackage{xcolor}
\usepackage{booktabs}
\usepackage{graphicx}
\usepackage{caption}
\usepackage{multirow}
\usepackage{amsfonts}
\usepackage{subcaption}

\newcommand{\Tref}[1]{Table~\ref{#1}}

\newcommand{\fref}[1]{Fig.~\ref{#1}}
\newcommand{\Fref}[1]{Figure~\ref{#1}}
\newcommand{\sref}[1]{Sec.~\ref{#1}}

\newcommand{\npeg}{N^\text{P}}
\newcommand{\nparticle}{K}
\newcommand{\mocopeg}{f_\text{PP}}
\newcommand{\mocohole}{f_\text{HP}}

\newcommand{\cam}[1]{#1}
\newcommand{\argmax}{\mathop{\rm arg~max}\limits}

\begin{document}
    \title{Tactile-Filter:\\ Interactive Tactile Perception for Part Mating}

    \vspace{-3mm}
    \author{
    \authorblockN{Kei Ota}
    \authorblockA{Mitsubishi Electric
    }
    \and
    \authorblockN{Devesh K. Jha}
    \authorblockA{Mitsubishi Electric Research Labs
    }
    \and
    \authorblockN{Hsiao-Yu Tung}
    \authorblockA{MIT}
    \and
    \authorblockN{Joshua B. Tenenbaum}
    \authorblockA{MIT}
    }
    


    \twocolumn[{%
    \renewcommand\twocolumn[1][]{#1}%
    \maketitle
    \begin{center}
        \centering
        \vspace{-10mm}
        \includegraphics[width=\textwidth]{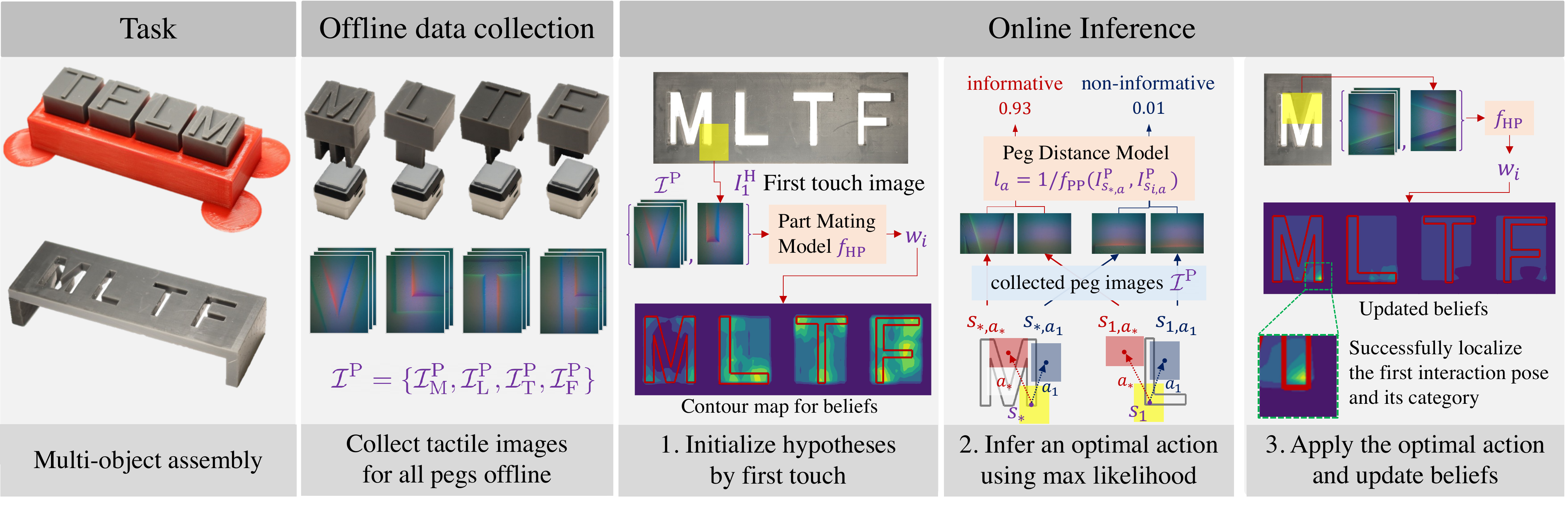}
        \captionof{figure}{
            Overview of the proposed \textit{Tactile-Filter}. As shown in the figure, we consider the task of the part mating without any prior knowledge of \cam{3D mesh of objects and} which objects fit together. We assume that the robot has access to a collection of tactile images for the set of pegs as shown in figure \cam{(Offline data collection)}. During inference, the robot tries to identify which peg would fit into a given hole by the proposed \textit{Tactile-Filter}. An initial set of hypotheses (denoted by $s \in \mathcal{S}$) is generated using the tactile image from the first touch and a trained part mating model, which predicts the correspondence between parts that fit together. We compute an optimal action for sampling the next image on the hole surface in order to minimize the uncertainty of the current estimate using a maximum likelihood approach. This is also illustrated in the figure, where given an initial touch, we can select an optimal action that results in maximum reduction of uncertainty. This method allows us to find the peg for the right fit as well as localize the hole (as we finally get the correct hypothesis) while minimizing the number of interactions during the task. (MLTF stands for Maximum Likelihood Tactile Filter). [Best viewed in color]
        }
        \label{fig:pipeline}
    \end{center}%
    }]
    \begin{abstract}
Humans rely on touch and tactile sensing for a lot of dexterous manipulation tasks. Our tactile sensing provides us with a lot of information regarding contact formations as well as geometric information about objects during any interaction. With this motivation, vision-based tactile sensors are being widely used for various robotic perception and control tasks. In this paper, we present a method for interactive perception using vision-based tactile sensors for a part mating task, where a robot can use tactile sensors and a feedback mechanism using a particle filter to incrementally improve its estimate of objects \cam{(pegs and holes)} that fit together.
To do this, we first train a deep neural network that makes use of tactile images to predict the probabilistic correspondence between arbitrarily shaped objects that fit together. The trained model is used to design a particle filter which is used twofold. First, given one partial (or non-unique) observation of the hole, it incrementally improves the estimate of the correct peg by sampling more tactile observations. Second, it selects the next action for the robot to sample the next touch (and thus image) which results in maximum uncertainty reduction to minimize the number of interactions during the perception task. We evaluate our method on several part-mating tasks \cam{with novel objects} using a robot equipped with a vision-based tactile sensor. We also show the efficiency of the proposed action selection method against a naive method. See supplementary video at \url{https://www.youtube.com/watch?v=jMVBg_e3gLw}.
\end{abstract}
    \IEEEpeerreviewmaketitle
    
    \section{Introduction}\label{sec:intro}
Humans rely on tactile sensing to monitor and control manipulation tasks during interactions with the environment. 
Tactile sensing allows us to monitor and respond to contact forces, adapt to object slip during grasping, and perform various perception tasks to build models of the environment. Even in situations where visual observation is not possible, tactile sensing enables us to interpret different types of interactions with the environment. For instance, we can locate objects in cluttered environments even in the absence of visual cues, identify the correct key for a lock by feeling the different options, or determine the type of video port (e.g., HDMI) on a monitor screen by touch alone. It has been the long-standing goal of robotics to imitate such intelligent behavior during manipulation tasks. Motivated by this goal, we present an interactive perception method for robots that utilize vision-based tactile sensors to construct reliable models for part mating. 

A lot of manufacturing tasks could be decomposed into a sequence of insertion tasks. Object insertion is a well-studied contact-rich manipulation task in robotics~\cite{dong2021tactile, siciliano2008springer}. However, the task becomes extremely challenging when the geometry of the mating objects is unknown. Also, this can make the task of part mating significantly complex as the uncertainty in geometry can limit the ability to understand any possible contact formation between the parts~\cite{dong2021tactile}. This complexity is further amplified in assembly tasks that require precise geometric information, as tolerances between mating parts become critical. Achieving the required level of precision in manufacturing tasks can be challenging when relying solely on vision-based algorithms.

\cam{
To address these challenges, we propose leveraging vision-based tactile sensors located at the robot's gripper for precise perception in these tasks. 
We present an interactive perception method, \textit{Tactile-Filter}, using vision-based tactile sensors for estimating part correspondence for part mating, i.e., to estimate which parts fit into each other using tactile sensors in the absence of any vision sensing. We train a deep learning model to predict correspondence between the correct mating parts, observed using a tactile sensor. In the presence of partial observation or non-unique contact patch, we make use of a particle filter to aggregate the information from multiple touches and improve the estimate of the correct fit for a given hole. To minimize the number of interactions between the robot and the object, a maximum likelihood-based action selection method is used during the proposed interactive perception. The proposed method is tested on several different test environments with objects of different shapes and sizes \cam{that are not used in training the model} to show the generalization of the proposed approach. \Fref{fig:pipeline} shows the abstract idea of the proposed interactive perception method.
}

\textbf{Contributions:} This paper has the following contributions:
\cam{
\begin{enumerate}
    \item We present a part mating problem that deals with objects of unknown shapes, aiming to identify and estimate the pose of mating parts using minimal number of interactions. To address this problem, we present a novel approach called \textit{Tactile-Filter}, which combines contrastive learning, particle filter, and tactile sensing for part mating.
    \item Through our experiments conducted on novel objects, we demonstrate that our proposed method effectively resolves uncertainty by iteratively updating its belief during interactions. We demonstrate the ability to generalize to objects not encountered during training. Furthermore, we introduce an action selection method within our approach, which leads to significant improvements in efficiency.
\end{enumerate}
}
It is noted that for the sake of brevity and clarity of presentation, we will use the word \textit{peg} for the male part and \textit{hole} for the female part in our paper.

    \section{Related Work}\label{sec:related_work}
Vision-based tactile sensors have attracted a lot of attention recently~\cite{yuan2017gelsight,taylor2022icra}. These sensors provide a high-resolution capture of the contact patch during contact formation and thus can help in localization of contacts, detection of slip, etc~\cite{taylor2022icra, ma2021extrinsic, dong2019maintaining}. Consequently, vision-based tactile sensors have been used for a lot of control and perception tasks~\cite{she2021cable, hogan2020tactiledexterity, dong2021tactile, shirai2023tactile, kim2023simultaneous}. Most of these methods make use of displacement of the contact patch to recover a signal indicating slip which can be used for control of the manipulation task. Similarly, there has been prior work that uses these sensors for object classification and slip stabilization using feedback from these sensors~\cite{calandra2018more, yuan2018active}, or for learning visuotactile servoing~\cite{chaudhury2022using, hensen2022visuotactilerl}.

\cam{
Several studies have focused on estimating object pose using tactile sensors which can be broadly categorized into two directions.
The first group of studies addresses pose estimation from a single touch~\cite{li2014localization,bauza2022tac2pose,fu2022safely,kelestemur2022tactile}, employing regression~\cite{li2014localization,fu2022safely} and contrastive learning~\cite{bauza2022tac2pose}. However, these methods exhibit limitations when applied to large objects or objects with non-unique contact patches.
The second group of studies focuses on pose estimation of large objects, often utilizing particle filtering techniques to narrow down the distribution of possible poses~\cite{luo2015localizing,kelestemur2022tactile,suresh2022midastouch,caddeo2023collisionaware}. However, these methods face challenges when applied to real objects due to factors such as low-dimensional sensors~\cite{luo2015localizing}, the requirement of a large number of interactions for training interaction policies~\cite{kelestemur2022tactile} or collecting samples~\cite{suresh2022midastouch}, and the reliance on 3D models of objects and/or simulators~\cite{kelestemur2022tactile,suresh2022midastouch,caddeo2023collisionaware}.
In contrast to these two sets of works, we assume the shape of the objects is not known a priori, and allow only a limited number of interactions (a maximum of $10$) by generating informative actions that effectively reduce ambiguity. Furthermore, our method not only estimates object pose but also identifies the object type from multiple candidates, adding an additional layer of complexity to the problem.}

\cam{There exists another category of works that integrate vision and tactile sensors to estimate object pose~\cite{luo2015localizing,snehal2022visuotactile,chaudhury2022using,fu2022safely}. By leveraging the capabilities of vision sensors, these methods can reduce the number of interactions required to estimate object pose by narrowing down the distribution of possible poses. While our study specifically focuses on object pose estimation using vision-based tactile sensors, our method is not mutually exclusive with the methods that combine vision and tactile sensors. By integrating both modalities, we can leverage the strengths of each sensor type and potentially improve the accuracy and efficiency of object pose estimation.}

Our problem setup is relevant to the interactive perception literature, where the goal is also to interact with the objects and update iteratively on the estimation of the states~\cite{bohg2017interactive,xia2020interactive,srivastave2022behavior}. While previous work focuses mostly on state estimation from raw visual perception~\cite{hausman2015active,bohg2017interactive}, point clouds~\cite{ota2022hsaur,wu2022vatmart,mo2021where}, or 3D states of the objects~\cite{novkovic2020object}. In contrast to these previous works, we consider state estimation using tactile sensors. 

Our work is also related to perception during insertion or part mating. Vision is mostly insufficient to perform a lot of insertion tasks due to the precision required during these tasks~\cite{dong2021tactile, lee2019making}.
Consequently, there has been a lot of work making use of tactile and/or wrench measurements on contact formation between the mating parts. The idea behind most of these tasks is to make use of tactile measurements and a feedback mechanism to iteratively correct the pose error between the mating parts~\cite{dong2021tactile, 8968204, 9838102}. In all these methods, geometric information is not explicitly used. Furthermore, they do not explicitly update the uncertainty in measurements. In contrast, we present a method where the robot can make use of the geometric information upon contact formation using tactile sensors to iteratively estimate the part correspondence as well as precise localization.

    \section{Problem Statement}\label{sec:problem_statement}
In this section, we present a formal statement of the problem which is studied in this paper. We also motivate the problem by discussing some common scenarios where the proposed problem could arise and the proposed method could be useful.

We consider the task of perception during part mating while performing automated assembly. We consider scenarios where the robot can not use a vision sensor to perceive the target object to perform the desired mating task. Such situations could arise in tasks where a robot has to assemble a product where occlusions are created by other parts (e.g., consider the assembly of an electronic board). Apart from occlusion, these tasks could also require precision in pose estimates which might be very difficult to obtain using vision. To formalize the problem, we define the task for the robot as identifying the correct peg from a known set of possible choices by multiple observations of the hole using the vision-based tactile sensor(s). The goal here is to design algorithms that can identify the correct peg with minimum physical interaction with the hole using a tactile sensor. We make the following assumptions in the proposed study:
\begin{enumerate}
    \item The possible number of pegs for the part mating task is fixed and known as apriori.
    \item The rough location of the target hole is known so that the robot can establish initial contact with the part and it does not need to perform this rough localization using touch.
    \item The robot can collect a dense set of tactile images for all the candidate pegs by touching each of the pegs at various different locations and orientations, prior to experiments.
\end{enumerate}
The first assumption is not restrictive as we will generally have a limited number of parts to assemble. The second assumption is also very easily met using common vision methods with rough precision in localization. The third assumption would require that the robot has access to a detailed geometric model of the possible pegs observed using tactile sensors. This would require that the robot performs some exploration to collect this data. This assumption is required as the size of the tactile sensor may be small compared to the size of the objects that the robot is interacting with.



We focus on the setup where the target objects are larger than the size of a tactile sensor such that the entire object can not be observed using a single touch or a single touch can result in non-unique observations (consider when parts of the objects could be similar). Such problems would require multiple touches and a method to aggregate the data from multiple touches.
For example, consider the task of inserting pegs into the shape of alphabet letters that are larger than the sensor size (see Fig.~\ref{fig:board_design}). If the target hole is the letter "A", whether we can be certain that it is indeed an "A" with a single touch depends on where we land our finger. For instance, if the first touch is made on the horizontal line segment of the letter "A," this feature may also be present in other alphabets such as "B," "D," "E," and so on. However, by making contact with features unique to the letter "A," such as the lower left intersection, the probability of other candidates can be reduced. So the question is how do we aggregate information from multiple touches, and how do we select places to touch that will maximize the information so as to reduce required interactions?

    \section{Tactile filter}

We propose \textit{Tactile-Filter}, an uncertainty-aware interactive perception method to identify the correct peg from a candidate set that fits into a given hole for assembly using tactile sensors.
At the core of the framework is a feature-matching model that computes the probability that a peg can pair with a hole by measuring the distance between the corresponding tactile images in a joint feature space. Using the feature matching model, we can construct the three critical steps in standard particle filtering: (1) we can initialize a set of possible hypotheses from the first touch by comparing the tactile image on the hole with the candidate images for the pegs collected before the experiment, (2) we then sample the most probable hypothesis and generate an action that maximizes uncertainty reduction, and (3) we apply the inferred action to the real system and update beliefs about the shape and pose of the target hole. Next, we detail each of these components.

\subsection{Learning to find mating part} \label{subsec:model}
Given a pair of peg and hole images, we train a model that maximizes the similarity score if the image for the hole corresponds to the image for the peg. To this end, we use a contrastive learning framework~\cite{chen2020simple,he2020momentum} to learn the feature space, similar to the work MoCo-v3~\cite{chen2021mocov3}.

MoCo-v3 has two encoders, $f_q$ and $f_k$, with output vectors $q$ and $k$. The goal of learning is to find the key vectors $k$ that correspond to the query vectors $q$. In our case, we consider the query vectors $q$ to be the vectors from images of the holes and learn to maximize the similarity score between $q$ and the vectors computed from the corresponding peg images $k$, while minimizing the similarity between the query vectors and a set of vectors from the negative peg images $\{ k^- \}$.
In MoCo-v3, this is formulated by minimizing the InfoNCE loss~\cite{oord2018representation}:
\begin{equation}
    \mathcal{L}=-\log \frac{\exp \left(q \cdot k^{+} / \tau\right)}{\exp \left(q \cdot k^{+} / \tau\right)+\sum_{k^{-}} \exp \left(q \cdot k^{-} / \tau\right)},
\end{equation}
where $\tau$ is the hyperparameter. More details can be found in~\cite{chen2021mocov3}, implementations in~\cite{mmselfsup2021}, and the training procedure is shown in \fref{fig:data_collection}.
We denote the model used for measuring similarities between the hole images and peg images as $\mocohole$, and name it \textit{part mating model}. 
Additionally, we train a model, denoted as $\mocopeg$ and referred to as the \textit{peg distance model}, to calculate the similarity between peg images. This model will be utilized to produce informative actions (as detailed in Section \ref{subsec:gen_act}).

\subsection{Generating hypotheses}
In cases where only a partial shape of a hole can be observed using a tactile sensor, multiple corresponding choices for peg may exist. Thus it might be hard or impossible to determine the object with a deterministic approach. To address this uncertainty, we explicitly generate and maintain a set of hypotheses representing potential candidate choices using a particle filter.

\begin{figure}[t]
    \centering
    \includegraphics[width=\columnwidth]{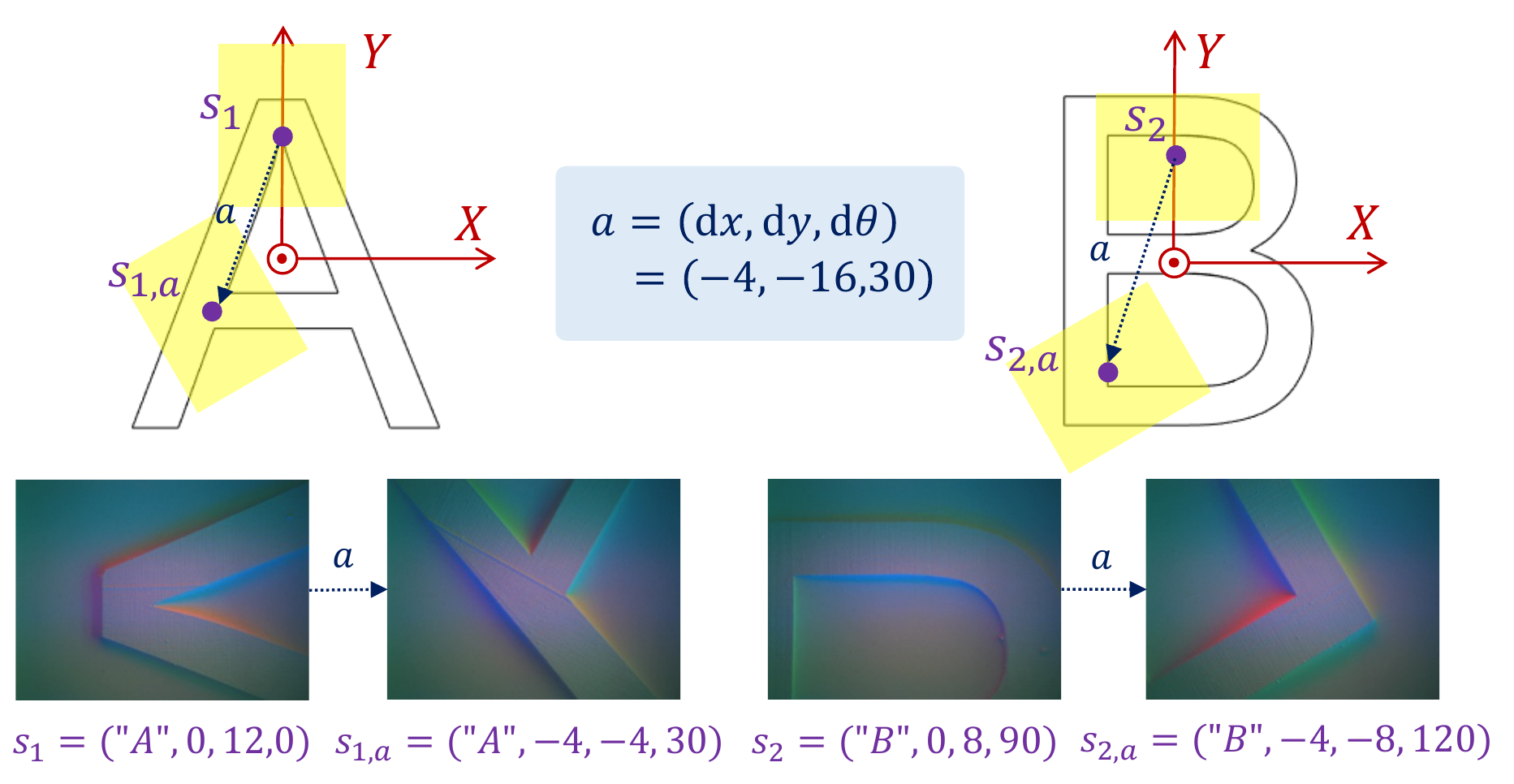}
    \caption{This picture defines particles and actions available to the robot. A particle $s$ is defined as a tuple consisting of the class of the object and a pose in SE(2) w.r.t. a frame attached to the center of the object. Each particle $s_i$ is associated with the corresponding image $I^\text{P}_{s_i}$ observed via the tactile sensor at that location. An action $a$ is equivalent to the transform in SE(2) applied to a particle $s_i$. The pose of a particle obtained by applying an action $a$ can be obtained by applying the transform in SE(2) to the pose of the particle $s$, and we denote it as $s_{i,a}$. Thus, an action $a$ will result in the observation of the contact patch $I^\text{P}_{s_{i,a}}$ at the new pose corresponding to the particle $s_{i,a}$. [Best viewed in color]}
    \label{fig:init_particles}
\end{figure}

Since the class of the target hole and its orientation and location is unknown (\sref{sec:problem_statement}), we generate a set of hypotheses $\mathcal{S}$ where each hypothesis $s_k$ is a quadruple: $s_k = (c_k, x_k, y_k, \theta_k)$, where $c_k$ represents the possible object categories $c_k \in \mathcal{C}$, and $x_k, y_k, \theta_k$ are the SE(2) relative planar displacements from the center of the object.
An example is shown in \fref{fig:pipeline} and is also explained in \fref{fig:init_particles}.

One can initialize the particle set by sampling from a uniform distribution within a reasonable range (e.g., $c_k$ from candidate categories, $x_k$ from $[-X_\text{MAX}, X_\text{MAX}]$, which is the lower and upper limits of reasonable sizes of the peg, etc.). However, it will be inefficient as the sampling space becomes huge as the target object becomes larger, and/or the number of candidate categories increases. As an alternative, we initialize the particle set after obtaining the first tactile image of the target hole $I_{t=1}^\text{H}$ (which we denote as $I_{1}^\text{H}$) by utilizing the previously collected set of peg images $\mathcal{I}^\text{P}$ and pre-trained part mating model $\mocohole$. Specifically, we compute the similarities, $w_i$, between the observed initial tactile image for the hole $I_{1}^\text{H}$ and each tactile image of the peg $I_i^\text{P}$ from the set of peg images $\mathcal{I}^\text{P}$ as:
\begin{equation}
    w_i = \mocohole (I_1^\text{H}, I_i^\text{P}). \label{eq:part_mating_computation}
\end{equation}
We sample a particle proportional to this likelihood. Therefore, the probability of a particle given the initial hole image can be written as:
\begin{equation}
    p(s = i | I_1^\text{H}) = \frac{w_i}{\sum_{i \in \{ 1, ..., |\mathcal{I}^P| \} }w_i} \label{eq:prior}
\end{equation}
where, $i$ is the index of the set of previously collected peg images as $i \in \{ 1, ..., | \mathcal{I^\text{P}} | \}$ (see \fref{fig:pipeline}).
We then initialize the set of hypotheses by independently sampling $\nparticle$ particles with the above distribution. It is noted that the above distribution is a categorical distribution over all peg images.

\begin{algorithm*}[tb]
\caption{\textit{Tactile-Filter}}
\label{alg:multi-touch-pf}
\hspace*{\algorithmicindent} \textbf{Input} Number of candidate pegs $\npeg$. A set of dense tactile images for each peg categories $\mathcal{I}^\text{P}=\{ \mathcal{I}_1^\text{P}, \cdots, \mathcal{I}_{\npeg}^\text{P} \}$, number of particles $\nparticle$, maximum number of interactions $N^\text{max}$, pre-trained part mating model $\mocohole$ and peg distance model $\mocopeg$ (\sref{subsec:model}), threshold to stop iteration $\delta^\text{prob}$\\
\hspace*{\algorithmicindent} \textbf{Output} Mating peg category and its displacement in $x,y,\theta$ coordinate from the center of the peg
\begin{algorithmic}[1]
    \State Touch and observe the first hole image $I_1^\text{H}$
    \State Compute similarity score $w_i$ between the hole image $I_1^\text{H}$ and a peg image $I_{i}^\text{P} \in \mathcal{I}^\text{P}$ as $w_{i} = f_\text{HP}(I_1^\text{H}, I_{i}^\text{P})$
    \State Initialize a particle pool $\mathcal{S} = \emptyset$
    \For{$k \gets 1$ to $\nparticle$}
        \State Sample a particle from the set of the peg images according to the categorical distribution defined in Eq.\eqref{eq:prior}
        \State Add the sampled particle $s_k = (c_k, x_k, y_k, \theta_k)$ to the particle pool: $\mathcal{S} \leftarrow \mathcal{S} \cup s_k$.
    \EndFor
    \For{$t \leftarrow 2$ to $N^\text{max}$}
        \State Select the most probable particle $s_\ast = \argmax_{s_k \in \mathcal{S}} w_k$
        \State Infer the optimal action $a_\ast$ according to Eq.~\eqref{eq:optimal_act} using $s_\ast$ and $\mathcal{S}$. 
        \State Move the robot with the inferred action $a_\ast$ and get the tactile image $I_t^\text{H}$
        \For{$k \leftarrow 1$ to $\nparticle$}
            \State Compute importance weight for the particle $w_k = \mocohole(I_t^\text{H}, I_{s_k}^\text{P})$
        \EndFor
        \State Compute the posterior distribution defined in Eq.\eqref{eq:posterior}
        \State Re-sample $K$ particles from $\mathcal{S}$ using the posterior distribution
        \State Compute updated posterior $p_t (c| I^\text{H}_{1:t})$ of each peg category from the particles as defined in Eq.~\eqref{eq:compute_class_prob}
        \If {$\max_{j\in \{1,...,\npeg \} } p_t(c_j) > \delta^\text{prob}$}
            \State Break the current loop
        \EndIf
    \EndFor
    \State Select the most probable particle $s_\ast = ( c_\ast, x_\ast, y_\ast, \theta_\ast) =\argmax_{s_k \in \mathcal{S} } w_k$ (to get the localization estimate, i.e., $x_\ast, y_\ast, \theta_\ast$)
    \State \Return Object category $c_\ast$ and its displacement $x_\ast, y_\ast, \theta_\ast$.
\end{algorithmic}
\end{algorithm*}

\subsection{Selecting informative action}\label{subsec:gen_act}
In order to efficiently determine the category and pose of the target hole, we aim to calculate an optimal action that can maximize uncertainty reduction. While it is possible to compute such an optimal action by maximizing information gain against all possible peg images, it necessitates the integration of all latent variables, making it computationally infeasible within a reasonable time frame. As an alternative, we utilize the existing hypothesis set $\mathcal{S}$ to enhance the sample efficiency.

We sample the most probable hypothesis from the current set of particles $s_\ast = \argmax_{s_k \in \mathcal{S}} w_k$ and determine the optimal action by simulating it on a set of previously collected peg images $\mathcal{I}^\text{P}$ (see \fref{fig:pipeline}). The action $a = (\mathrm{d}x, \mathrm{d}y, \mathrm{d}\theta)$ is represented as a transform in SE(2) to the pose of the particles, and we denote the particle $s_k$ with the updated pose by applying the action $a$ as $s_{k,a}$.
With this updated pose and the peg images $\mathcal{I}^\text{P}$, we can also obtain the peg image when applied to the action $a$, which we denote $I_{s_{k, a}}^\text{P}$. If there is no corresponding pose in the collected set of images, we assign an empty image which is a tactile image without any contact $I_\text{NoContact}$. We visually explain the definition of particles and actions in~\fref{fig:init_particles}.

Given a current most-likely hypothesis, the next optimal action can be selected by finding the most informative action. To do that, we compute the distance between the tactile images obtained by applying any possible action to the most probable hypothesis and the remaining particles in set $\mathcal{S}$. The action that maximizes the sum of this distance over all the particles is selected as the optimal action.
Such an action is favored only when the peg image of the sampled particle is close to the hole image, while other images have a greater distance when applying the same action to all the other particles. More concretely, we define the likelihood of an action as
\begin{equation}
    l_a = \sum_{s_i \in \mathcal{S} \setminus \{ s_\ast \} } 1 / \mocopeg (I_{s_{\ast,a}}^\text{P}, I^\text{P}_{s_{i,a}}). \label{eq:peg_distance_computation}
\end{equation}
The optimal action can be then selected by maximizing the likelihood as:
\begin{equation}
    a^\ast = \argmax_{a \in \mathcal{A}}  l_a, \label{eq:optimal_act}
\end{equation}
where $\mathcal{A}$ is a set of actions with which the tactile sensor can observe the peg after applying the action. 
This action set $\mathcal{A}$ can be obtained by calculating the L1 pixel distance between the tactile image of the peg after applying the action $I_{s_{\ast,a}}^\text{P}$ and the peg image that does not have contact $I_\text{NoContact}^\text{P}$, and see if it exceeds a threshold as $\mathbb{I}( \| I_{s_{\ast,a}}^\text{P} - I_\text{NoContact} \|_1 < \delta^\text{act} )$.

\cam{
This procedure is visually depicted in~\fref{fig:pipeline}, where the optimal action $a^\ast$ is shown to generate a more distinct contact patch (observable through touch) when applied to all particles ($s_\ast$ and $s_1$ in the figure). On the other hand, the non-informative action $a_1$ produces similar contact patches that would not effectively disambiguate the current belief.
}

\subsection{Update hypotheses}
After obtaining the optimal action $a_t$ at time step $t$, we apply it on the real robot and observe the tactile image of the hole $I_{t}^\text{H}$. We then update the probability of each hypothesis by comparing the observed hole image $I_{t}^\text{H}$ and the peg images from the current hypotheses $I_{s_{k,a_t}}^\text{P}$:
\begin{equation}
    \begin{split}
        p(s_t | I_{1:t}^\text{H}, a_{1:t-1})
        & \propto p(I_t^\text{H} | s_{t}) p(s_t|I_{1:t-1}^\text{H}, a_{1:t-1})\\
        & \approx \sum_{k=1}^K w_k p(s_t|I_{1:t-1}^\text{H}, a_{1:t-1}),
        \label{eq:posterior}
    \end{split}
\end{equation}
where $w_k = \frac{\mocohole (I_t^\text{H}, I_{s_{k,a_t}}^\text{P})}{\sum_{k=1}^K \mocohole (I_t^\text{H}, I_{s_{k,a_t}}^\text{P})} $ is the likelihood (weight) for the hole image to match with the peg image of the $k^{th}$ particle.
The second term can be written as:
\begin{equation}
    \begin{split}
        p(s_t | I^H_{1:t-1}
        &, a_{1:t-1}) \\ 
        &= \sum_{s_{t-1}\in \mathcal{S}} \underbrace{p(s_t | s_{t-1}, a_{t-1})}_{\text{forward dynamics}} \underbrace{p(s_{t-1} |  I^H_{1:t-1}, a_{1:t-1})}_{\text{obtain through recursion}},
    \end{split}
    \label{eq:pred}
\end{equation}
where the first term represents the deterministic forward dynamics. Given the particle is in state $s_k$ at time step $t$, the dynamics can be expressed as:
\begin{equation}
p(s_{t+1} | s_{t}, a_{t}) = \left\{
\begin{array}{ll}
1 & \text{if } s_{t+1} = s_{k,a_t}\\
0 & \text{otherwise}.
\end{array}
\right.
\end{equation}
The second term in Eq.\eqref{eq:pred} is initialized with the prior defined in Eq.~\eqref{eq:prior} and can be obtained through recursion.
We update the distribution of the particles by regenerating a new set of particles through weighted sampling based on $w$.
The full algorithm is shown in Alg.~\ref{alg:multi-touch-pf}.

\begin{figure*}[t]
    \centering
    \includegraphics[width=0.99\textwidth]{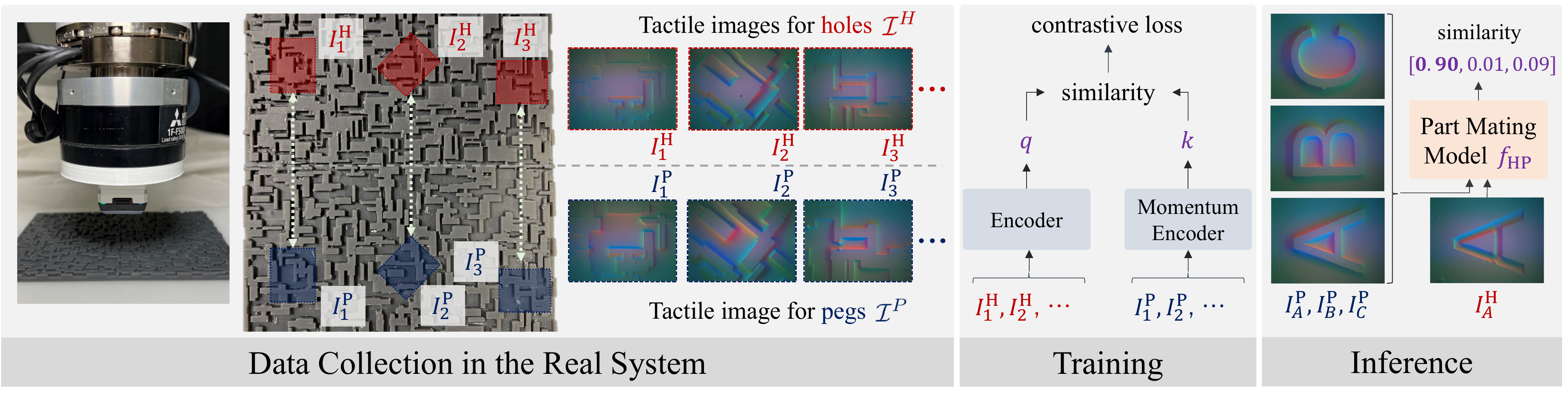}
    \caption{Data Collection, Training, and Inference of the Part Mating Model: The left block depicts the data collection process using the \textit{MAZE} board that features various shapes, including hole and peg shapes in the upper and lower halves, respectively. This board is placed on the robot platform and the robot arm equipped with a tactile sensor at the tip of the wrist makes contact with the board to collect data denoted as $\mathcal{I}^\text{H}$ and $\mathcal{I}^\text{P}$, which corresponds a set of images for pegs and holes, respectively. The middle block illustrates the training procedure for the part mating model. It is trained in a self-supervised manner using a contrastive loss that encourages the model to produce high scores only when images corresponding to true mating parts are provided. The right block demonstrates the model's generalization to different shapes after training. [Best viewed in color]}
    \label{fig:data_collection}
\end{figure*}

\cam{
\subsection{Terminal condition}
After updating the posterior of the particles with Eq.~\eqref{eq:posterior}, we compute the posterior probability of each peg category as follows:
\begin{equation}
    p_t (c|I_{1:t}^H) = \frac{\sum_{s_k \in \mathcal{S}} \mathbb{I}(s_k \in c)}{|\mathcal{S} |}, \label{eq:compute_class_prob}
\end{equation}
where $\mathbb{I}(s_k \in c)$ is an indicator function that returns $1$ only when the category of the particle $c_k$ belongs to the category $c$. The algorithm terminates when the majority of particles belong to a specific class, indicated by $\max_{j\in \{1,...,\npeg \} } p_t(c_j) > \delta^\text{prob}$, which is a user-specified parameter for termination.
}
    \section{Experiments}
In this section, we evaluate the performance of the \textit{Tactile-Filter} algorithm in two different test scenarios. The first scenario, referred to as the \textbf{small objects}, involves a collection of small objects that can be fully captured by a single touch of the tactile sensor, thereby making the estimation problem relatively simpler to solve. The second scenario, referred to as the \textbf{large objects}, involves objects that are larger than the size of the tactile sensor, requiring multiple touch measurements to accurately estimate their shape. \cam{All the test objects used in the pose estimation experiments are novel and are not used for training the contrastive learning model.}

\subsection{Training the part mating model}

\textbf{Tactile sensor.}
We use a commercially available GelSight Mini~\cite{gelsightmini2023} tactile sensor, which provides $320 \times 240$ compressed RGB images through the Robot Operating System (ROS) at a rate of approximately 25 Hz, with a field of view of $18.6 \times 14.3$ millimeters.

\textbf{Robot platform.}
The MELFA ASSISTA robot~\cite{assista}, a collaborative robot with 6 DoF, is used in this study. The tactile sensor is mounted on the robot's wrist during data collection (see \fref{fig:data_collection}). It is noted that we do not use the force torque sensor mounted at the wrist of the robot as shown in the \fref{fig:data_collection}.

\textbf{Data collection.}
In order to train a model that is capable of generalizing to a diverse set of shapes, we designed a board for data collection so that it features random polygonal shapes to simulate pegs and holes of arbitrary shapes. The shapes were generated through a process that involved creating a maze (we name it \textit{MAZE} board), adding random perturbations to the position and size of the walls that make up the maze, and then exporting the result for 3D printing. This board was designed such that any arbitrary hole patch sampled from the upper half has a corresponding mating peg patch in the lower half (see \fref{fig:data_collection}). To collect data for training, we sampled several different locations and orientations on the upper half MAZE board from a high-resolution grid to collect the hole images, and then collect the corresponding peg images from the lower half. This resulted in a total of approximately $23,000$ pairs of images of pegs and holes which perfectly fit with each other.

\textbf{Preprocessing.}
In this study, the tactile sensor used has RGB LEDs with different colors on each of the three surfaces~\cite{gelsightmini2023}. As a result, even when the same object is in contact, the color may differ depending on the position of the image captured. To mitigate the potential impact on generalization performance, we obtained an image of a non-contact situation $I_\text{NoContact}$ during data collection, reducing the impact by subtracting the image. Then, the average and variance of each RGB channel were calculated for all images, and the images were normalized before being input into the model.

\textbf{Training.}
As described in~\sref{subsec:model}, we use MoCo-v3 for our part mating model $\mocohole$ and peg distance model $\mocopeg$. We train the models with the collected images using the MAZE board for $500$ epochs. To improve generalization capability, we augment the data by using random cropping and horizontal or vertical flips, which will be applied to the pairs of images inputted to the model during training.

\subsection{Small objects} \label{subsec:exp_small_obj}
We first evaluate the performance of the TactileFilter when applied to objects that fit in the size of the sensor.

\textbf{Baselines.}
To understand the challenges encountered when identifying objects that might not be fully captured through a single touch, we compare our method against two methods that only use the initial image. The first baseline, referred to as \textit{Pixel}, computes the L1 distance between the peg and hole images and returns the index of the nearest neighbor image.
The second baseline, \textit{MoCo}, utilizes the pre-trained MoCo-v3 model to calculate the distance (negative of the MoCo-v3's output) based solely on the first tactile image and without incorporating any subsequent interactions.
The results of our method are denoted as \textit{Ours ($n$)}, where $n$ indicates the number of interactions. It is important to note that the value of $n$ includes the initial contact, therefore, \textit{Ours ($n=1$)} represents the results obtained without any additional interactions.

\begin{figure}[t]
    \centering
    \includegraphics[width=\columnwidth]{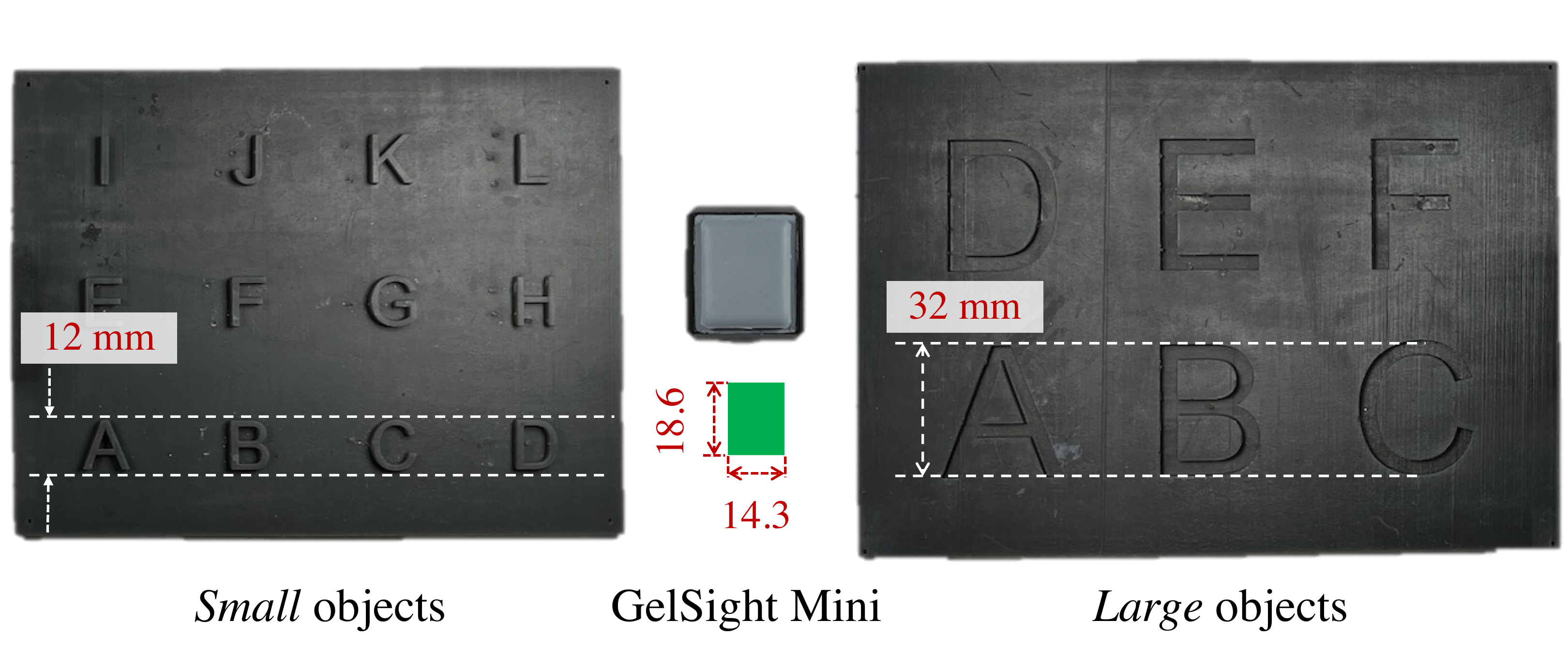}
    \caption{Alphabet boards for our experiments. The left board contains small characters, each with a length of $12$ mm and a maximum width of $16$ mm, to fit within the size of the sensor if the robot makes contact with the center position. The sensor size is shown in the middle image. The board on the right has large characters with a length of $32$ mm and a maximum width of $40$ mm, requiring multiple interactions with the tactile sensor to obtain complete geometry for the object.}
    \label{fig:board_design}
\end{figure}

\textbf{Settings.}
For this experiment, we have designed an evaluation board consisting of \cam{$12$ alphabet characters (ranging from “A” to “L”)}, each with a maximum width of $16$ mm and height of $12$ mm, so the characters fit within a single touch. 
Since we would like to evaluate the model in situations where the pose of the object is unknown, resulting in only partial observation of the object and requiring multiple touches for accurate estimations, we collect data with displacements in $X, Y \in \{ -8, -4, 0, 4, 8 \}$ millimeter and $\theta \in \{ -90, -60, ..., 90 \}$ degree from their center position. This results in \cam{$12 \times 5 \times 5 \times 7 = 2100$} images.
\Fref{fig:board_design} shows examples of the characters we used for the experiment. The hyperparameter we used for our algorithm is the number of particles $\nparticle=100$, the maximum number of iterations $N^\text{max}=10$, and the threshold to stop the iteration $\delta^\text{prob}=0.95$.

\textbf{Metrics.}
The performance of the results is assessed through two metrics. Firstly, we evaluate the accuracy in classifying the objects. For the baseline calculation, we calculate the distance between a hole image and all previously gathered peg images, select the image with the minimum distance, and consider the prediction to be accurate when the predicted image's class matches the class of the inputted hole image. Additionally, the distance between the predicted pose and the ground truth pose is quantitatively measured.

With regard to the evaluation of the proposed method, we utilize the likelihood used for updating the particles to weight the prediction. The object with the highest weighted probability is then evaluated with the target object. We also use weighted error between the particles and the ground truth image to compute the quantitative error.

\textbf{Results and Analysis.}
The results are presented in \Tref{tab:results_alphabet_single}. A comparison between the two baselines, \textit{Pixel} and \textit{MoCo}, reveals that correspondences between parts cannot be obtained simply by comparing pixel values. The contrastive framework captures the features of mating parts, resulting in improved performance. However, the results using only the first contact are still not sufficiently accurate as the tactile sensor only observes a partial view of the object. In contrast, our method demonstrates a gradual improvement in performance as interactions are added. Additionally, as we can see from Table~\ref{tab:results_alphabet_single}, our method is able to achieve good localization accuracy both in position and orientation. In particular, we are able to achieve \cam{a submillimeter average error} in localization which might be required for industrial insertion tasks.

\begin{table}[t]
    \centering
    \caption{Quantitative evaluation of single touch experiments with \textbf{small} objects on the alphabet board.}
    \begin{tabular}{l|rrrrr}
        \toprule
        & \multirow{2}{*}{Pixel} & \multirow{2}{*}{MoCo} & \multicolumn{3}{c}{Ours} \\ 
        & & & $n=3$ & $n=5$ & $n=10$\\ \midrule
        \cam{Accuracy [\%]} & $0.0$ & $39.6$ & $81.8$ & $90.7$ & $95.0$ \\
        Error $XY$ [mm] & - & $0.7$ & $0.2$ & $0.1$ & $0.1$ \\
        Error $\theta$ [deg] & - & $5.4$ & $0.9$ & $0.3$ & $0.1$ \\
        \bottomrule
    \end{tabular}
    \label{tab:results_alphabet_single}
\end{table}

\begin{table}[t]
    \centering
    \caption{Quantitative evaluation of multiple touch experiments with \textbf{large} objects on the alphabet board.}
    \begin{tabular}{c|rrrrr}
        \toprule
        & \multirow{2}{*}{Pixel} & \multirow{2}{*}{MoCo} & \multicolumn{3}{c}{Ours} \\ 
        & & & $n=3$ & $n=5$ & $n=10$\\ \midrule
        Accuracy [\%] & $11.9$ & $41.9$ & $58.7$ & $72.3$ & $85.0$ \\
        Error $XY$ [mm] & $6.9$ & $4.4$ & $1.3$ & $1.0$ & $0.7$\\
        Error $\theta$ [deg] & $16.6$ & $15.5$ & $4.4$ & $2.9$ & $1.5$\\
        \bottomrule
    \end{tabular}
    \label{tab:results_alphabet_multi}
\end{table}

\begin{figure*}[t]
\begin{minipage}{0.33\linewidth}
    \centering
    \includegraphics[width=0.95\textwidth]{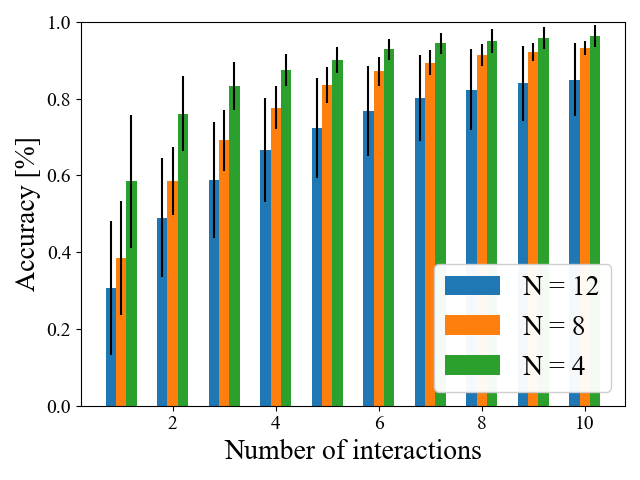}
    \caption{Classification accuracy of the proposed method with a different number of classes evaluated on the \textit{Large} objects. The result shows our method can quickly identify the correct class if the number of classes (shown by $N$) is limited.}\label{fig:different_classes}
\end{minipage}
\begin{minipage}{0.02\linewidth}
\end{minipage}
\begin{minipage}{0.65\linewidth}
    \begin{minipage}{0.49\linewidth}
        \centering
        \includegraphics[width=0.95\textwidth]{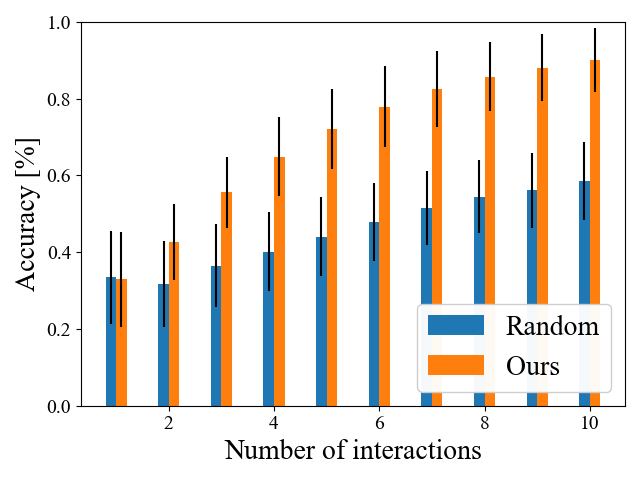}
        \subcaption{\textit{Small} objects}\label{fig:results_small_ablation}
    \end{minipage}
    \begin{minipage}{0.49\linewidth}
        \centering
        \includegraphics[width=0.95\textwidth]{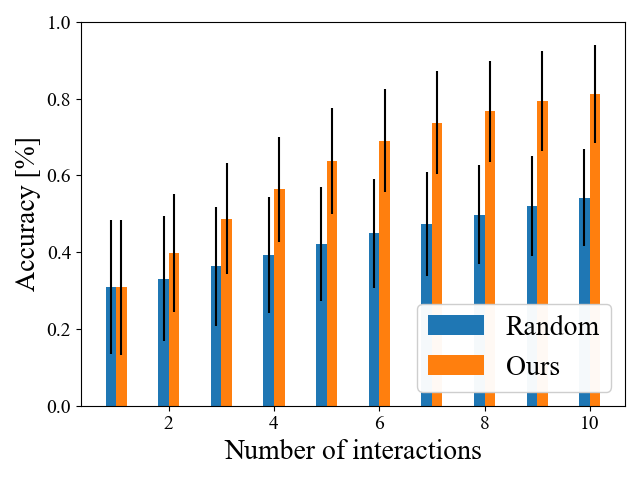}
        \subcaption{\textit{Large} objects}\label{fig:results_large_ablation}
    \end{minipage}
    \centering
    \caption{Classification accuracy for \cam{\textit{Small}} and \textit{Large} objects with different action strategies. As could be seen from these bar plots, our proposed method demonstrates significant improvement in comparison to the random action selection with regard to classification accuracy.}
    \label{fig:ablations_action_strategy}
\end{minipage}
\end{figure*}

\subsection{Large objects}
\textbf{Settings.} In the next set of experiments, we evaluate the performance of the proposed method when applied to objects that are larger than the size of the sensor. This scenario requires the robot to interact multiple times with the object to gain a comprehensive understanding of its shape. To this end, we have designed an evaluation board consisting of twelve alphabet characters (ranging from “A” to “L”), each with a maximum width of $40$ mm and height of $32$ mm.
We tested the method on the location and orientation of the robot from $X, Y \in \{ -20, -16, ..., 20 \}$ mm and $\theta \in \{ -90, -60, ..., 90 \}$ with respect to the center position of each character.
\Fref{fig:board_design} presents examples of the characters utilized in the experimental setup.
As for the baselines, we compare the method against the same baselines as the previous experiment on \textit{small} objects.

\textbf{Results and Analysis.}
\Tref{tab:results_alphabet_multi} shows the results on the large objects. Similar to the results obtained in the setting of small objects, our proposed model demonstrates improved performance compared to the baselines. However, it is also observed that a larger object size requires a greater number of interactions in order to achieve comparable accuracy.
In \Fref{fig:different_classes}, we show the classification accuracy with respect to the number of interactions, and the results with randomly sampled smaller sets of $4$ and $8$ characters to evaluate the performance with a smaller number of possible candidates. The bar plots demonstrate that the proposed method can quickly identify the correct class if the number of classes is small.

\subsection{Ablation on action selection strategy}
\textbf{Settings.}
To assess the effectiveness of the proposed action selection strategy, we compare the proposed method with a random action selection method (which we call \textit{Random}). We evaluate the two methods on the \textit{Small} and \textit{Large} objects settings described earlier.

\textbf{Results and Analysis.}
The results in \fref{fig:ablations_action_strategy} indicate the proposed maximum likelihood action selection approach demonstrates significant improvement in comparison to the method with regard to classification accuracy.

\subsection{Evaluation on industrial connectors}
\begin{figure}[t]
    \centering
    \includegraphics[width=0.99\columnwidth]{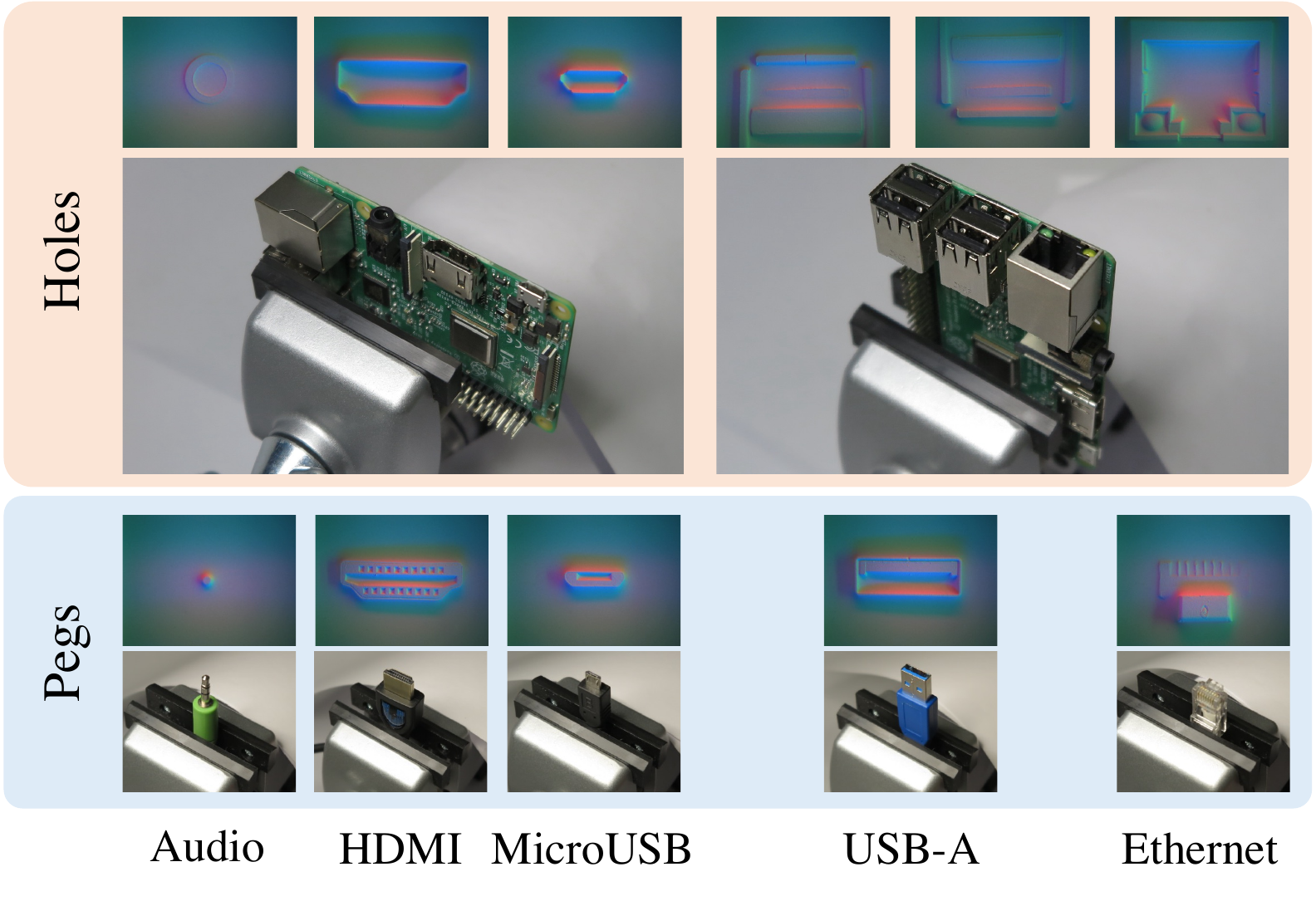}
    \caption{Experimental setup for industrial connector identification on a Raspberry-Pi board. This image shows the  observations of the six pegs and holes using the GelSight Mini sensor. Table~\ref{tab:results_raspi} shows the classification results obtained by our model. [Best viewed in color]}
    \label{fig:raspi_setting}
\end{figure}

\textbf{Settings.}
To further evaluate the performance of the trained part mating model in an industrial setting, we collect tactile images of connectors and sockets from a Raspberry-Pi board, as depicted in \fref{fig:raspi_setting}.

\textbf{Results and Analysis.}
The results of the evaluation of the Pixel baseline and our part mating model for the classification of connectors and sockets from the Raspberry-Pi board are presented in \Tref{tab:results_raspi}. The Pixel baseline demonstrates improved performance in comparison to the small and large object experiments, due to the reduced number of classes in this setting and the unique size of each connector/socket, which simplifies the classification through the use of only L1 pixel distance. Although the part mating model outperforms the Pixel baseline, it misclassifies the female HDMI connector as the male USB-A connector. This is attributed to the significant distribution shift between the training set and the test set, where the pins on the surface of the male part are not present in the training data. To address this issue, future work can focus on enhancing the generalization capabilities of the part mating model.

\begin{table}[t]
    \centering
    \caption{Classification accuracy on the Raspberry Pi Board.}
    \begin{tabular}{c|ccc}
        \toprule
         & Pixel & MoCo \\
        Accuracy [\%] & $50.0$ & $83.3$ \\
        \bottomrule
    \end{tabular}
    \label{tab:results_raspi}
\end{table}

\begin{figure*}[t]
\begin{minipage}{0.49\linewidth}
    \centering
    \includegraphics[width=\columnwidth]{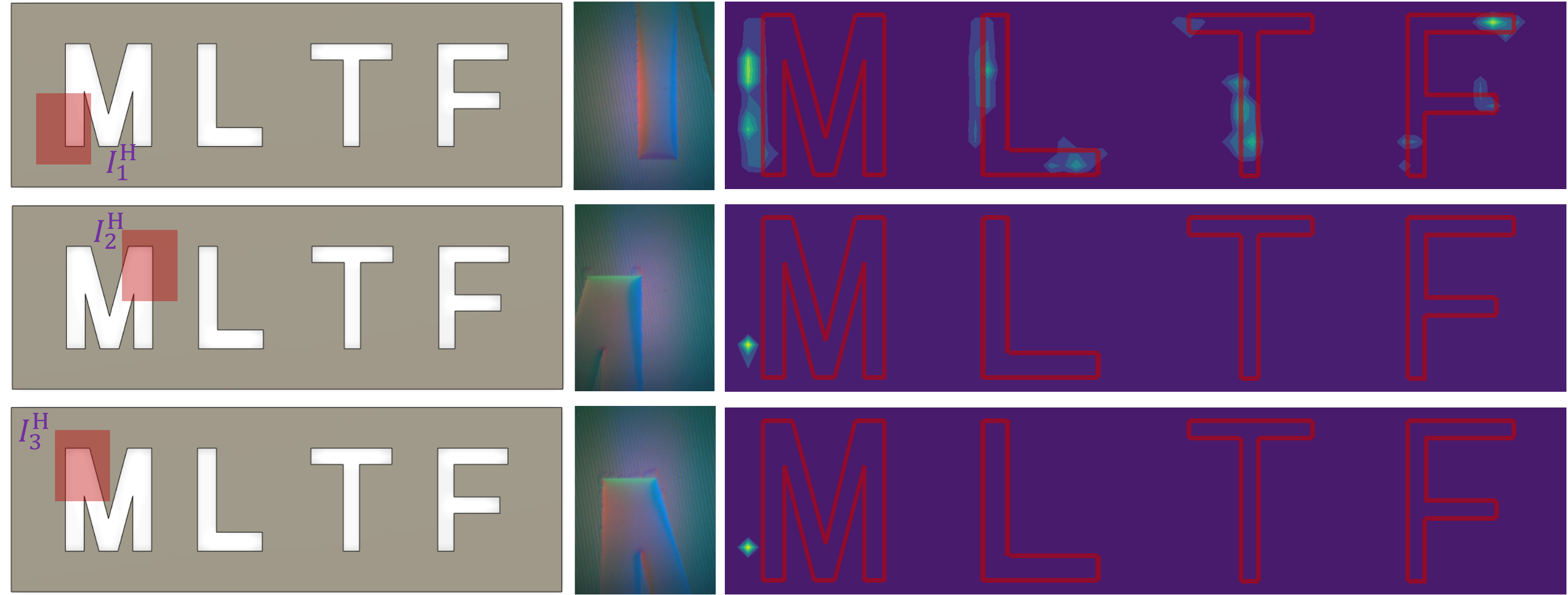}
    \subcaption{``M''}
    \label{fig:mltf_m}
\end{minipage}
\begin{minipage}{0.49\linewidth}
    \centering
    \includegraphics[width=\columnwidth]{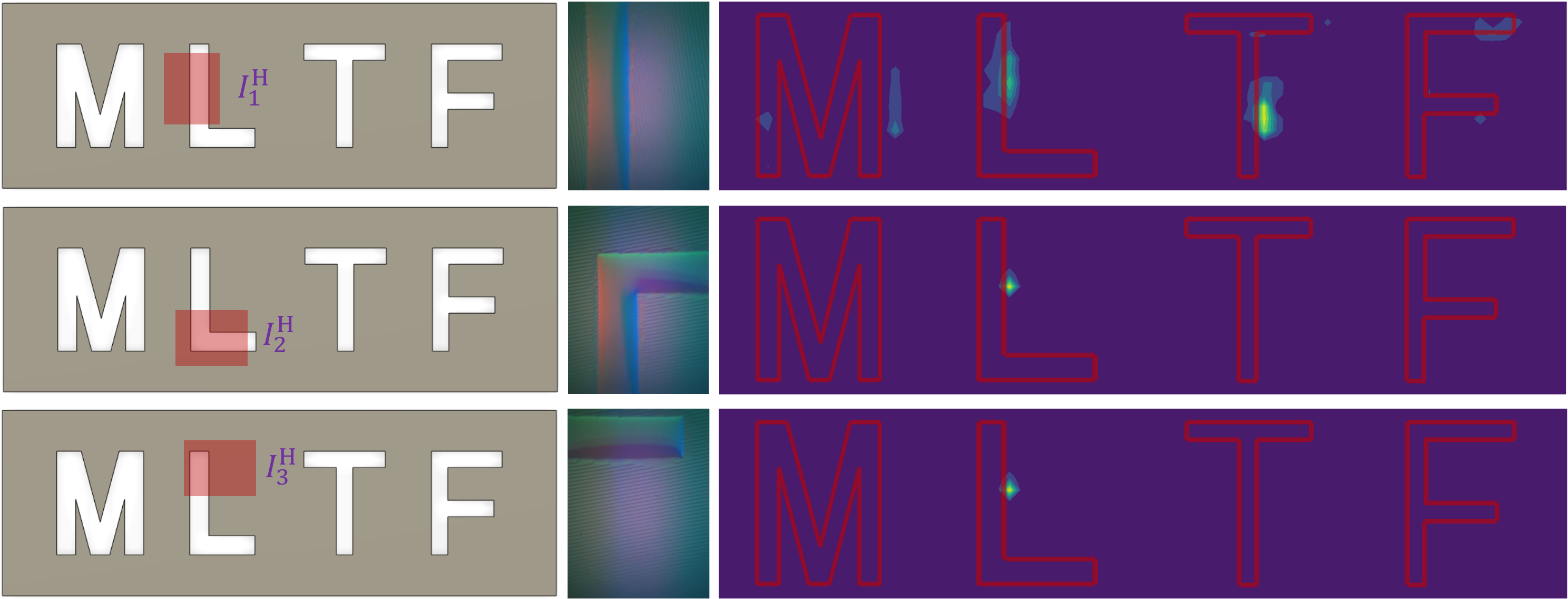}
    \subcaption{``L''}
    \label{fig:mltf_m}
\end{minipage}\\
\begin{minipage}{0.49\linewidth}
    \centering
    \includegraphics[width=\columnwidth]{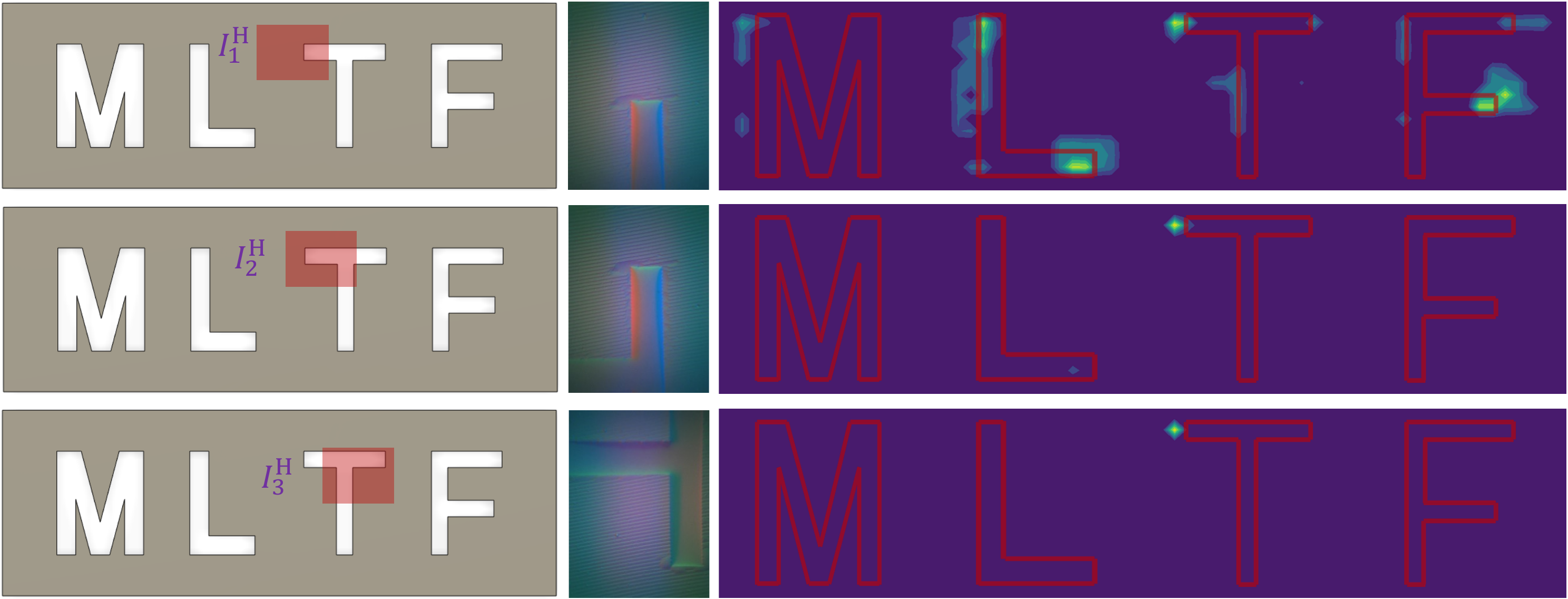}
    \subcaption{``T''}
    \label{fig:mltf_t}
\end{minipage}
\begin{minipage}{0.49\linewidth}
    \centering
    \includegraphics[width=\columnwidth]{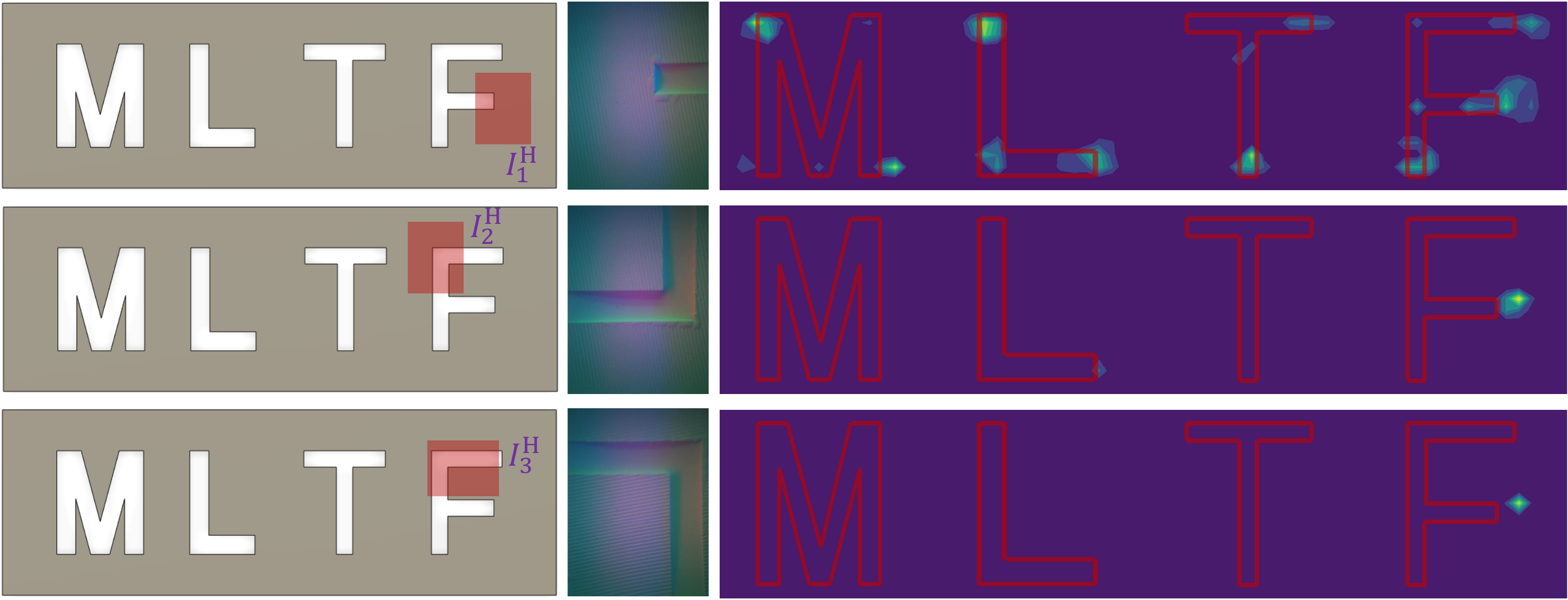}
    \subcaption{``F''}
    \label{fig:mltf_f}
\end{minipage}
\caption{\cam{Visualization of belief maps for holes categorized as ``M'', ``L'', ``T'', and ``F''. In each figure, the red regions in the left column show the spatial and temporal region captured by the tactile sensor during interactions, represented as $I_t^\text{H}$, where $t$ indicates the interaction number or timestep. The center image shows the hole image captured by the tactile sensor, and the right figure illustrates the belief map generated by the current particles. Across all hole types, the results indicate a rapid convergence of the distribution of the hole's initial contact pose and its corresponding category.}}
\label{fig:mltf_visualizations}
\end{figure*}

\cam{
\subsection{Application to multi-object assembly}
\textbf{Settings.}
Once Tactile-Filter localizes the hole and identifies the corresponding peg, the robot can successfully insert the peg into the right hole. This is facilitated by the algorithm's ability to estimate the pose with high precision, as demonstrated by the submillimeter average prediction error (refer to \Tref{tab:results_alphabet_single} and \Tref{tab:results_alphabet_multi}). Consequently, we assess the proposed method in a real multi-object assembly scenario during the final experiment.}

\cam{
The task involves identifying and localizing four pegs and holes shaped like the alphabet characters “M”, “L”, “T”, and “F” (Maximum Likelihood Tactile Filter). Each peg has a length of $32$ mm and a maximum width of $28$ mm, as depicted in \fref{fig:pipeline} (leftmost picture). Although the algorithm achieves accurate pose estimation for the holes, the robot still fails in insertion due to errors during grasping. To account for this, we design holes with a tolerance of 2 mm and treat insertion as a simple pick-and-place operation using an impedance controller. For more precise assembly, we can combine our method with prior work, such as \cite{dong2021tactile,9561646}.}

\cam{
\textbf{Results and Analysis.} 
The qualitative results are available in the video accessible at \url{https://www.youtube.com/watch?v=jMVBg_e3gLw}. The video demonstrates the algorithm successfully identifies the correct peg and pose of the hole, and the robot successfully inserts the pegs. Moreover, the visualization generated from this experiment as shown in \fref{fig:mltf_visualizations} demonstrates that our method is iteratively corrects its belief during the interactive perception. Finally, in terms of computational time, the most time-consuming steps of the algorithm involve updating importance weights using the part mating model $\mocohole$ in Eq.~\eqref{eq:part_mating_computation} and the peg distance model $\mocopeg$ in Eq.~\eqref{eq:peg_distance_computation}. However, each of these steps takes approximately 0.3 seconds on a single GPU, which is significantly shorter than other robot operations. Thus, our algorithm is suitable for online control.
}
    \section{Conclusions and Future Work}\label{sec:conculsion}
Tactile sensing can allow robots to build reliable models of their environment to perform precise manipulation tasks. In this paper, we presented a novel method called \textit{Tactile-Filter}. We presented an interactive perception method where a robot can improve its estimate for the perception task using tactile sensors while minimizing the number of interactions required with its environment. We considered the design of the method in the context of the task of part mating. In the absence of any vision input, we described a method where the robot could incrementally improve its estimate of correspondence between parts within a fixed number of available choices. We also proposed a maximum likelihood-based approach to select future actions to minimize the number of interactions during the perception task. The proposed method was verified using a vision-based tactile sensor and a physical robot on several tasks of part mating. The generalization of the proposed method to previously unseen scenarios was also illustrated.

As our method was trained and evaluated on a physical system, data collection was performed using a real robot. However, in future work, we aim to explore the possibility of utilizing an appropriate simulation environment~\cite{church2021optical, xu2022efficient, Xu-RSS-21, wang2022tacto} to simulate contact patches for various object geometries. This approach would enable us to reduce reliance on physical robots for data collection and instead leverage simulations to acquire data. By doing so, we can learn and develop complex perception techniques with minimal usage of real-world data.

Another limitation of the proposed method is that we assume that tactile images consist of only the peg and hole parts. Therefore, the underlying deep learning model can get easily confused if distractors are present in the image (such as attachments to connectors in \fref{fig:raspi_setting}). In future research, we can work on this limitation by localizing the mating parts from the tactile image to make the method more robust to such distractors.

    \bibliographystyle{plainnat}
    \bibliography{references}
\end{document}